\documentclass[conf]{new-aiaa}
\usepackage[utf8]{inputenc}
\usepackage[T1]{fontenc}
\usepackage{graphicx}
\usepackage{amsmath}
\usepackage[version=4]{mhchem}
\usepackage{siunitx}
\usepackage{longtable,tabularx}
\usepackage{ragged2e}
\usepackage{colortbl}
\usepackage{float}
\usepackage{multicol}
\usepackage{algorithm} 
\usepackage{algpseudocode} 
\usepackage{adjustbox}

\title{Synthesizing Follow-Up Drive Data for Enhanced Road Safety in Intelligent Driving Function Systems}

\author{\textbf{Nico Schick}\footnote{N. Schick, M. Sc. studied Applied Computer Sciences (M. Sc.) and Computer Engineering (B. Eng.) at the Esslingen University of Applied Sciences. e-mail: Nico.Schick@hs-esslingen.de}, Franjo Čičak}

\begin{document}

\maketitle

This study underscores the vital importance of intelligent driving functions in enhancing road safety and driving comfort. Central to our research is the challenge of obtaining sufficient test data for evaluating these functions, especially in high-risk, safety-critical driving scenarios. Such scenarios often suffer from a dearth of available data, primarily due to their inherent complexity and the risks involved.

Addressing this gap, our research introduces a novel methodology designed to create a wide array of diverse and realistic safety-critical driving scenarios. This approach significantly broadens the testing spectrum for driver assistance systems and autonomous vehicle functions. We particularly focus on the follow-up drive scenario due to its high relevance in practical applications. Here, vehicle movements are intricately modeled using kinematic equations, incorporating factors like driver reaction times. We vary parameters to generate a spectrum of plausible driving scenarios.

The utilization of the Difference Space Stopping (DSS) metric is a pivotal element in our research. This metric plays a crucial role in the safety evaluation of follow-up drives, facilitating a more thorough and comprehensive validation process. By doing so, our methodology enhances the reliability and safety assessment of driver assistance and autonomous driving systems, specifically tailored for the most challenging and safety-critical scenarios.

\section{Motivation}
The integration of intelligent driving functions represents a transformative step in advancing road safety and driver comfort, particularly in safety-critical scenarios. These scenarios, where the need for flawless and error-free system performance is paramount, highlight the critical role of reliable autonomous driving technologies \cite{FahrerAssSysImproveSafe}.

A primary challenge in this field is acquiring comprehensive test data from these critical scenarios. The presence of inherent risks and intricacies in safety-critical situations leads to a notable scarcity of accessible data, especially when compared to more typical driving scenarios. The scarcity of data in this context constitutes a significant obstacle when it comes to rigorously assessing and validating the performance of advanced driving functions.

Therefore, there's a need for innovative methodologies that can effectively expand the dataset for safety-critical scenarios, especially for testing and validation purposes. Comprehensive and rigorous testing is essential to enhance confidence in these intelligent systems, which in turn is crucial for ensuring the highest standards of road safety. Our research addresses this requirement by investigating innovative methods to improve and augment the quality of test data. This effort significantly contributes to the development and reliability of intelligent driving functions.

\section{Objective}
\label{sect:ziel}
The primary goal of this research is to develop a sophisticated methodology that significantly broadens the testing range of safety-critical driving scenarios for autonomous vehicles. 

This methodology addresses the prevalent challenge of the scarcity of safety-critical driving data, focusing on scenarios that can be represented as multivariate time series. The effectiveness and applicability of the generated driving scenarios are fundamentally dependent on their reflection of real-world driving dynamics and time-related characteristics.

A key aspect of our approach is the detailed analysis of a follow-up drive involving two vehicles. In this approach, we model the relative movements of these vehicles using kinematic equations, translating them into multivariate time series. A crucial consideration in our model is the inclusion of drivers' reaction times, ensuring a realistic representation of on-road dynamics. Through strategic adjustments of parameters within realistic driving dynamics, we achieve the generation of time series for vehicle movements that are not only plausible but also diverse.

The safety implications of these synthesized movements or time series are assessed using the Difference Space Stopping (DSS) metric as a foundation for this assessment. This capability enables us to effectively differentiate between safety-critical and non-critical driving scenarios. Consequently, our approach facilitates a more comprehensive validation of driver assistance systems and autonomous vehicle functions, ensuring extensive test coverage across a multitude of driving scenarios, varying environmental conditions, and potential edge cases.

Ultimately, our objective is to enhance the reliability and safety of driver assistance and autonomous vehicle systems through a methodical expansion of test data, thereby contributing significantly to the advancement of road safety standards.

\section{Time Series}
Time series are essential in data analysis, representing sequences of observations for specific characteristics tracked over time. These sequences are denoted as $t = t_1, \ldots, t_N$, where $N$ is the total number of observations. Time series data can be categorized into two types based on the data structure. Univariate time series consist of data points that are uniformly single-dimensional for each time instance. In contrast, multivariate time series involve data points that are multi-dimensional or vectors, with varying dimensions corresponding to each time stamp.

The nature of a time series is further distinguished by its domain – continuous or discrete. Continuous time series represent an uninterrupted flow of data, exemplified by constant sensor readings without gaps \cite{BAEU}. Discrete time series, in contrast, represent data collected at non-sequential time values, typical in event-driven scenarios.

Our research primarily relies on multivariate time series due to their direct relevance in real-world applications. In our study, the focus is on various driving scenarios, taking into account kinematic parameters such as position, speed, and acceleration, all aligned with a unified time vector. Formally, a multivariate time series can be defined as follows:
\begin{align}
f : D \in \mathbb{R} \mapsto Z \in \mathbb{R}^k, t \mapsto {Y_{t=t_1}^k, Y_{t=t_2}^k, \ldots, Y_{t=t_N}^k }, \#(t,Y)=N, k \in \mathbb{N}^+
\end{align}
In this regard, the time vector $t$ is synchronized with the sets of observations $Y^k$, representing various characteristics $k$. The dimensionality and complexity of these observations are represented by $N$ data points within the time series \cite{BARC}.

\section{Safety-critical Driving Scenarios}
\label{SafCritDrvScenSec}
Driving scenarios\footnote{Although \textit{driving situations} and \textit{driving scenes} are related terms, they are distinct from \textit{driving scenarios}. \textit{Driving situations} refer to specific drives with consistent characteristics such as traffic conditions and environmental factors \cite{DefFahrsituation}, whereas \textit{driving scenes} represent specific moments within these drives \cite{DefFahrszene}.} are characterized by dynamic demands of driving behavior, encompassing elements such as high-speed maneuvers, braking, cornering, and overtaking \cite{DefFahrszenario}. These scenarios can involve several vehicles, different road participants, and various environmental factors.

Assessing the importance of safety-critical factors within these driving scenarios is a complex and pivotal aspect of vehicular safety. The diversity of metrics and safety indicators in literature \cite{SafIndRef2} reflects the challenge of such assessments. The evaluation of scenario criticality can be approached both qualitatively and quantitatively. However, qualitative methods often face difficulties in establishing safety distinctions due to factors that can vary, such as kinematic parameters.

\begin{figure}[H]
\centering
\includegraphics[scale=0.6]{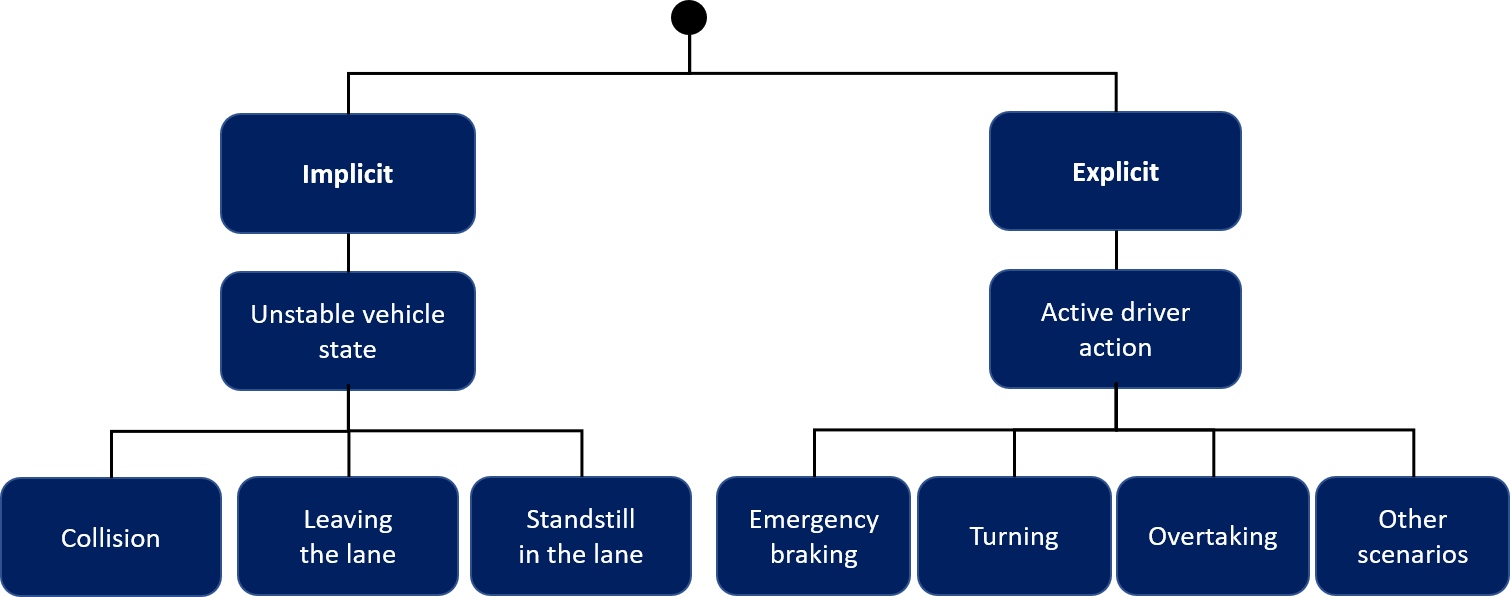}
\caption{Taxonomy of safety-critical driving scenarios}
\label{TaxoSafeDrvScen}
\end{figure}

It is particularly crucial to differentiate between safety-critical and non-critical driving scenarios. Safety-critical situations have substantial implications for the safety of all road users. Figure \ref{TaxoSafeDrvScen} offers a taxonomic view of these scenarios, illustrating their complexity and diversity. This implies that safety-critical driving situations are generally divided into two categories: implicit and explicit.

\subsection{Implicit Scenarios}
Implicit scenarios arise from unstable vehicle states, leading to lane deviations, unintended stops, or collisions. These can occur autonomously, driven by external factors such as environmental conditions or internal factors like human errors, underscoring the unpredictable nature of driving environments. Implicit scenarios frequently pose challenges to autonomous vehicle predictive algorithms, demanding a high level of situational awareness and rapid response mechanisms.

\subsection{Explicit Scenarios}
On the other hand, explicit safety-critical scenarios involve direct driver interventions. Notable instances, as discussed in \cite{B1}, encompass overtaking and turning maneuvers, as well as drives that entail reactive vehicle braking. These scenarios underscore the crucial role of driver responses in managing potentially risky situations.

\subsection{Challenges and Opportunities}
The diversity of driving scenarios is theoretically boundless, presenting a significant challenge in their categorization and distinction. This vast spectrum of scenarios, while enriching the range of possible situations, also complicates the identification and analysis of unique scenarios.

Data for safety-critical driving scenarios are notably scarce \cite{ChallengeWenigeDatenCritDriveScene2} \cite{ChallengeWenigeDatenCritDriveScene3}, a challenge emphasized in \cite{ChallengeWenigeDatenCritDriveScene1} due to the heightened restrictions on data availability compared to non-critical scenarios. Factors contributing to this scarcity include the rarity of such events, the need for standardized data collection, privacy concerns, and liability issues \cite{ReasonsWenigeDatenCritDriveScene}.

Despite these obstacles, the available data on safety-critical scenarios are invaluable. They provide insights into a broad spectrum of hazardous situations encountered in the real world. Integrating these scenarios into system testing enhances the robustness and reliability of autonomous driving systems. This inclusion broadens test coverage, aiding in the identification of potential system vulnerabilities and thereby strengthening the safety and effectiveness of autonomous vehicles.

\section{Follow-Up Drive}
The criticality of integrating safety-critical driving scenarios into the verification and validation of driving functions, as discussed in Section \ref{SafCritDrvScenSec}, is further emphasized in this section. We specifically focus on a follow-up drive involving two vehicles as a representative example of such scenarios.

\begin{figure}[H]
\centering
\includegraphics[scale=0.4]{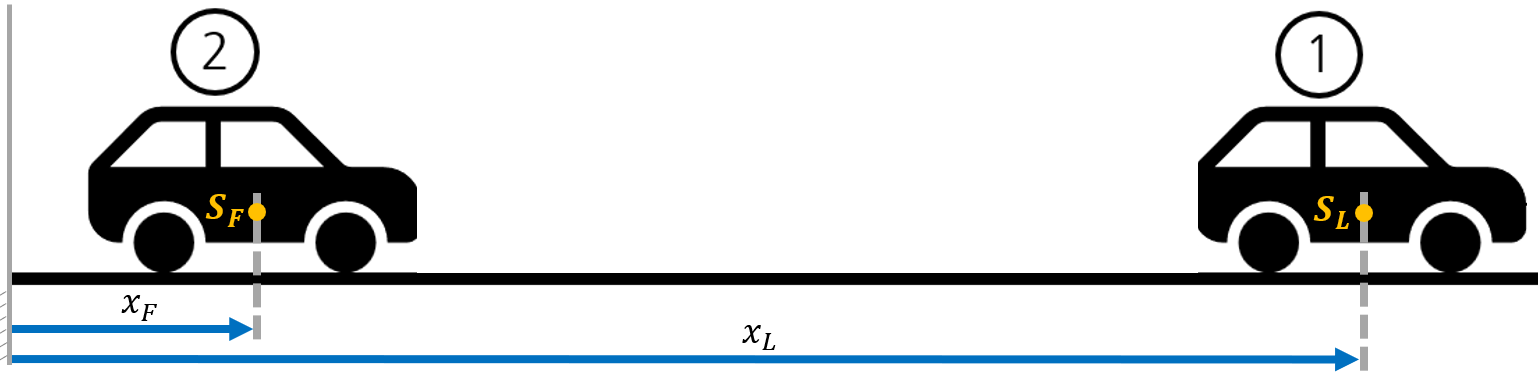}
\caption{Follow-Up Drive: Illustration}
\label{ValidScene}
\end{figure}

We selected the follow-up drive scenario for our analysis due to its frequent occurrence in real-world driving, its simplicity, and its relevance for modeling essential safety factors. A comprehensive analysis of this particular scenario is presented within this publication, accompanied by a visual representation of the follow-up drive's configuration in Figure \ref{ValidScene}.

\subsection{Illustration}
In this scenario, two vehicles are in motion along a longitudinal path, with Vehicle 2 following Vehicle 1. A significant aspect of this scenario involves the synchronized braking maneuvers executed by both vehicles at specified time. The kinematic parameters under examination include position ($x$), speed ($v$), acceleration ($a$), time ($t$), and the drivers' reaction times ($t_R$). While the relevance of jerk ($j = \dot{a}$) holds theoretical importance, our analysis primarily concentrates on the macroscopic motion of the vehicles rather than microscopic perspective.

\newpage

\subsection{Dynamics of Vehicle Motion}
\label{sec:dynamicsVehicleMotion}
The dynamic interaction between the vehicles is governed by their respective equations of motion. These equations are pivotal in providing a detailed understanding of the vehicles' kinematic behaviors as they evolve over time, a fundamental aspect for analyzing the intricacies of vehicular dynamics on the road. Below, we will provide a more in-depth explanation of the motion equations for each individual vehicle.

\subsubsection{Kinematics of the Leading Vehicle (L)}
The velocity \(v_{L}(t)\) of the leading vehicle \(L\) at a specific time \(t\) is influenced by its initial velocity \(v_{0,L}\), the initial acceleration \(a_{0,L}\), and the elapsed time \(t - t_{R,L}\) since the beginning of the observation. The term \(t_{R,L}\) is incorporated to reflect the driver's reaction time for vehicle \(L\), accounting for the delay in driver's ability to adjust the velocity of vehicle \(L\). 
\begin{equation}
    v_L(t) = \begin{cases}
        v_{0,L} & t \leq t_{R,L} \\
        v_{0,L} + a_{0,L} \, (t-t_{R,L}) & t > t_{R,L}
    \end{cases}
\end{equation}
Similarly, the position \(x_{L}(t)\) of the leading vehicle \(L\) at a specific time \(t\) is a function of its starting position \(x_{0,L}\), initial velocity \(v_{0,L}\), initial acceleration \(a_{0,L}\), and the time difference \(t - t_{R,L}\). Here, \(t_{R,L}\) also serves to represent the driver’s reaction time for vehicle \(L\), influencing the positional changes of vehicle \(L\) over time.
\begin{equation}
    x_L(t) = \begin{cases}
        x_{0,L} + v_{0,L} \, t & t \leq t_{R,L} \\
        x_{0,L} + v_{0,L} \, t + \dfrac{1}{2} \, a_{0,L} \, (t-t_{R,L})^2  & t > t_{R,L}
    \end{cases}
\end{equation}

\subsubsection{Kinematics of the Trailing Vehicle (F)}
The velocity \(v_{F}(t)\) of the trailing vehicle \(F\) at a specific time \(t\) is influenced by its initial velocity \(v_{0,F}\), the initial acceleration \(a_{0,F}\), and the elapsed time \(t - t_{R,F}\) since the beginning. The term \(t_{R,F}\) is incorporated to reflect the driver's reaction time for vehicle \(F\), accounting for the delay in driver's ability to adjust the velocity of vehicle \(F\).
\begin{equation}
    v_F(t) = \begin{cases}
        v_{0,F} & t \leq t_{R,F} \\
        v_{0,F} + a_{0,F} \, (t-t_{R,F}) & t > t_{R,F}
    \end{cases}
\end{equation}
Similarly, the position \(x_{F}(t)\) of the trailing vehicle \(F\) at a specific time \(t\) is a function of its starting position \(x_{0,F}\), initial velocity \(v_{0,F}\), the initial acceleration \(a_{0,F}\), and the time difference \(t - t_{R,F}\). Here, \(t_{R,F}\) also serves to represent the driver’s reaction time in vehicle \(F\), influencing the positional changes of vehicle \(F\) over time.
\begin{equation}
    x_F(t) = \begin{cases}
        x_{0,F} + v_{0,F} \, t & t \leq t_{R,F} \\
        x_{0,F} + v_{0,F} \, t + \dfrac{1}{2} \, a_{0,F} \, (t-t_{R,F})^2  & t > t_{R,F}
    \end{cases}
\end{equation}

\newpage
\subsubsection{Effective Distance between Leading Vehicles (L) and Following Vehicle (F)}
\begin{figure}[H]
\centering
    \includegraphics[scale=0.6]{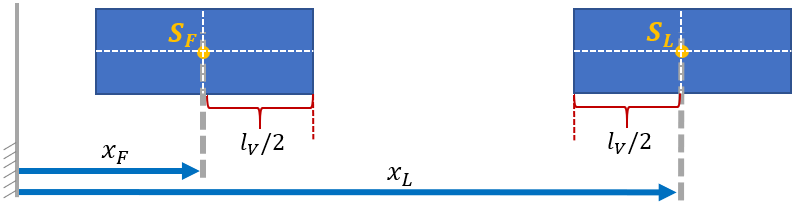}
    \caption{Follow-Up Drive: Importance of vehicle length}
    \label{IH}
\end{figure}

The distance between two vehicles depends on their positions and vehicle length ($l_V$), as shown in Figure \ref{IH}, where the vehicle length is assumed to be equal for both.
It is crucial to emphasize that Equation \ref{eq:motion5_xDiff} assumes a uniform mass distribution within both vehicles. This assumption results in the center of gravity, denoted as $S$, being positioned at the midpoint of each vehicle. Consequently, the safety-related effective distance between both vehicles, denoted as \(x_{Diff}(t)\), is defined as follows:
\begin{align}
    x_{Diff} = x_L - x_F - l_{V}
    \label{eq:motion5_xDiff}
\end{align}

\subsubsection{Driver's Reaction Time}
The reaction time $t_R$ is a significant parameter in safety-critical driving scenarios. Further information can be found in \cite{B1}. In experimental psychology, reaction time is defined as the duration from a stimulus to the corresponding response of an individual. Therefore, reaction times depend on psychological and physiological factors. In the field of vehicle movements, the driver's reaction time is also a crucial parameter, especially during braking maneuvers. \cite{B1}

The reaction time cannot be considered to be constant. It is strongly influenced by the criticality of the underlying situation, driver assessment, as well as the physical and mental state of the driver. 

Alternatively, reaction time can be estimated using a probability distribution function to incorporate the mentioned dependencies into a modeling of driving behavior. With respect to statistical analysis, the reaction time $t_R$ can be specifically modeled as a gamma distribution $\Gamma$ depending on the two scaling parameters $a$ and $b$ with $p$ as probability value. \cite{B1}

Below, we present the mathematical modeling of the reaction time $t_R$ using the gamma distribution and the two scaling parameters $a$ and $b$ \cite{matlabGamma} \cite{matlabGammaDist}:
\begin{align}
	t_R &:= F^{-1} (p|a,b) = \left\{ t_R : F(t_R|a,b) = p \right\}, \; a>0, \, b>0 \\
 	p &:= F(t_R|a,b) = \dfrac{1}{b^a \, \Gamma(a)} \, \int_{0}^{t_R} t^{a-1} \, e^{-t/b} \, dt \\
	\Gamma(a) &:= \int_{0}^{\infty} e^{-t} \, t^{a-1} \, dt
\end{align}
In this context, key statistical values regarding reaction times can also be derived. Specifically, the arithmetic mean, as an approximation of the expected value, is approximately 0.7 seconds. The standard deviation is approximately 0.2 seconds. The minimum and maximum reaction times are limited to 0.3 and 1.7 seconds, respectively. \cite{B1}

\subsection{Algorithm}
\label{sec:algorithmEqMotion}
To generate synthetic test data, especially multivariate time series for follow-up drives that depend on the motion equations elaborated in section \ref{sec:dynamicsVehicleMotion}, it is essential to utilize an appropriate algorithm. In the subsequent section, we present and elaborate on Algorithm \ref{algo:pseudo1}.

\begin{algorithm}[H]
  \caption{Pseudo code: Calculation of time series representing follow-up driving scenarios}
  \begin{algorithmic}[1]
    \For{$j := 1,\hdots,n-1$} \Comment{Iteration over all data points per time series}
      \State $t_j \gets t_{j-1} + \Delta t$ \Comment{Determination of the time vector used for all time series}
    \EndFor
    \For{$i := 0,\hdots,N-1$} \Comment{Iteration over all time series}  
      \State $a_{0,L}^{(i)} \gets \mathcal{N}(\mu_{a,L}, \sigma_{a,L}^2)$ \Comment{Initial acceleration per time series / Vehicle (L)}
      \State $a_{0,F}^{(i)} \gets \mathcal{N}(\mu_{a,F}, \sigma_{a,F}^2)$ \Comment{Initial acceleration per time series / Vehicle (F)}
      \State $t_{R,L}^{(i)} \gets F^{-1}(p\,|\,a,b)$ \Comment{Reaction time per time series / Vehicle (L)}
      \State $t_{R,F}^{(i)} \gets F^{-1}(p\,|\,a,b)$ \Comment{Reaction time per time series / Vehicle (F)}
    \EndFor
    \For{$i := 0,\hdots,N-1$} \Comment{Iteration over all time series}
        \For{$j := 0,\hdots,n-1$} \Comment{Iteration over all data points per time series}
        \If{$t^{(j)} \leq t_{R,L}^{(i)}$} \Comment{Reaction time not yet elapsed / Vehicle (L)}
            \State $v_L^{(i,j)} \gets v_{0,L}^{(i)}$ \Comment{Velocity / Vehicle (L)}
            \State $x_L^{(i,j)} \gets x_{0,L}^{(i)} + v_{0,L}^{(i)}\,t^{(j)}$ \Comment{Distance traveled / Vehicle (L)}
        \Else
            \State $v_L^{(i,j)} \gets v_{0,L}^{(i)} + a_{0,L}^{(i)}\left(t^{(j)}-t_{R,L}^{(i)}\right)$ \Comment{Velocity / Vehicle (L)}
            \State $x_L^{(i,j)} \gets x_{0,L}^{(i)} + v_{0,L}^{(i)} \, t^{(j)} + \dfrac{1}{2}\,a_{0,L}^{(i)}\left(t^{(j)}-t_{R,L}^{(i)}\right)^2$ \Comment{Distance traveled / Vehicle (L)}
        \EndIf
        \If{$t^{(j)} \leq t_{R,F}^{(i)}$} \Comment{Reaction time not yet elapsed / Vehicle (F)}
            \State $v_F^{(i,j)} \gets v_{0,F}^{(i)}$ \Comment{Velocity / Vehicle (F)}
            \State $x_F^{(i,j)} \gets x_{0,F}^{(i)} + v_{0,F}^{(i)}\,t^{(j)}$ \Comment{Distance traveled / Vehicle (F)}
        \Else
            \State $v_F^{(i,j)} \gets v_{0,F}^{(i)} + a_{0,F}^{(i)}\left(t^{(j)}-t_{R,F}^{(i)}\right)$ \Comment{Velocity / Vehicle (F)}
            \State $x_F^{(i,j)} \gets x_{0,F}^{(i)} + v_{0,F}^{(i)} \, t^{(j)} + \dfrac{1}{2}\,a_{0,F}^{(i)}\left(t^{(j)}-t_{R,F}^{(i)}\right)^2$ \Comment{Distance traveled / Vehicle (F)}
        \EndIf
        \EndFor
    \EndFor
  \end{algorithmic}
  \label{algo:pseudo1}
\end{algorithm}
\newpage
This algorithm is designed to calculate time series data representing follow-up driving scenarios, focusing on two vehicles: a leading vehicle (L) and a following vehicle (F).
\begin{enumerate}
    \item \textbf{Uniform Time Vector}: The algorithm first establishes a uniform time vector for all time series. It iterates over the desired data points, indexed by \(j\), ranging from 1 to \(n-1\). The time for each point, \(t_j\), is calculated by the sum of the previous time value, \(t_{j-1}\), and time step, \(\Delta t\).
    \item \textbf{Initialization}: The algorithm iterates over all the time series, each indexed by \(i\), from 0 to \(N-1\). For each time series, it initializes various parameters for both the leading and following vehicles:
        \begin{itemize}
            \item The initial accelerations \(a_{0,L}^{(i)}\) and \(a_{0,F}^{(i)}\) are generated from normal distributions with means \(\mu_{a,L}\) and \(\mu_{a,F}\), and variances \(\sigma_{a,L}^2\) and \(\sigma_{a,F}^2\) respectively.
            \item The reaction times \(t_{R,L}^{(i)}\) and \(t_{R,F}^{(i)}\) for each vehicle are determined using the inverse \(F^{-1}\) of the gamma distribution function depending on parameters \(p\), \(a\), and \(b\).
        \end{itemize}
    \item \textbf{Determine Vehicle Motion Values}: For each individual time series, the algorithm proceeds to iterate through all of its data points. During this iteration, it considers each data point within each time series:
        \begin{itemize}
            \item The algorithm verifies whether the driver's reaction time has passed for each vehicle. If it hasn't, the vehicle retains its initial velocity.
            \item Once the driver's reaction time has passed, the algorithm computes the updated velocity and the distance traveled by each vehicle. This calculation considers factors such as the initial velocity, initial acceleration, and the time that has elapsed since the driver's reaction time.
        \end{itemize}
\end{enumerate}

\section{Safety Indicator Metrics}
Safety indicator metrics play a crucial role in evaluating the severity of driving scenarios. Especially when examining longitudinal driving data presented in the form of time series, the literature referenced in \cite{B2} strongly recommend the utilization of the Difference Space Stopping (DSS) metric for evaluating follow-up drives.

\subsection{Difference Space Stopping (DSS)}
DSS serves as a safety indicator metric for evaluating follow-up driving scenarios. It is defined by calculating the difference between the spatial distance and stopping distance of two vehicles following each other. The spatial distance is obtained by summing the braking distance $x_{B,L}$ of the leading vehicle and the effective distance $x_{Diff}$ between the leading and following vehicles. The stopping distance is calculated by summing the brake reaction distance $x_{R,F}$ and the braking distance $x_{B,F}$ of the following vehicle. 

DSS can be understood as the static state of both vehicles when the leading vehicle is braking, subsequently leading the following vehicle to brake as well. To incorporate the worst-case braking effects into DSS, the maximum deceleration value $a_{min} = \mu \, g$ is applied for braking. \cite{DSS}
\begin{align}
DSS = \left(x_{Diff} + x_{B,L}\right) - \left(x_{R,F} + x_{B,F}\right) = \left( \left( x_L-x_F-l_V \right) + \frac{v_{L}^2}{2 \, a_{min}} \right) - \left( v_F \, t_R + \frac{v_{F}^2}{2 \, a_{min}} \right)
\end{align}

\subsection{Algorithm}
\label{sec:algorithmDSS}
To incorporate safety assessment through DSS evaluation for time series representing follow-up driving scenarios, it is essential to utilize an appropriate algorithm, as demonstrated in Algorithm \ref{algo:pseudo2}.
\begin{algorithm}[H]
  \caption{Pseudo code: Safety assessment through DSS evaluation for time series representing follow-up driving scenarios}
  \begin{algorithmic}[1]
    \For{$i := 0$ to $N-1$} \Comment{Iteration over all time series}
        \State $b_{C,1st} \gets \text{False}$ \Comment{Semaphore for the first safety-critical occurrence per time series}
        \For{$j := 0$ to $n-1$} \Comment{Iteration over all data points per time series}
            \If{$a_L^{(i)} < 0 \And a_F^{(i)} < 0$} \Comment{DSS calculation only in case vehicles are braking}
                \State $DSS^{(i,j)} \gets \left(\left(x_L^{(i,j)}-x_F^{(i,j)}-l_V\right) + \dfrac{v_L^{{2}\;(i,j)}}{2 \, a_{min}}\right)-\left(v_F^{(i,j)} \, t_{R,F}^{(i)} + \dfrac{v_F^{{2}\;(i,j)}}{2 \, a_{min}} \right)$ \Comment{DSS metric}
                \If{$DSS^{(i,j)} < 0$} \Comment{Data point of the time series with DSS value less than zero}
                    \State $t_C^{(i,j)} \gets t^{(j)}$ \Comment{Safety-critical time value}
                    \If{not $b_{C,1st}$} \Comment{Semaphore: First safety-critical occurrence per time series?}
                        \State $t_{C,1st}^{(i)} \gets t^{(j)}$ \Comment{First safety-critical time value per time series}
                        \State $b_{C,1st} \gets \text{True}$ \Comment{First safety-critical occurrence per time series identified}
                    \EndIf
                \EndIf
            \Else
                \State $DSS^{(i,j)} \gets \text{NaN}$ \Comment{No valid DSS calculation}
            \EndIf
        \EndFor
    \EndFor
  \end{algorithmic}
  \label{algo:pseudo2}
\end{algorithm}
\begin{enumerate}
\item \textbf{Time Series Iteration}: The algorithm iterates through each time series in the dataset, indexed by $i$, ranging from 0 to $N-1$, where $N$ is the total number of time series. For each time series, a flag $b_{C,1st}$ is initialized as False. This flag serves as a semaphore to mark the first occurrence of a safety-critical event within the time series.
\item \textbf{Data Point Iteration}: For each time series, the algorithm iterates over all its data points, indexed by $j$, ranging from 0 to $n-1$, where $n$ is the number of data points in the time series.
\item \textbf{Safety Assessment Condition}: The DSS metric is calculated only if both vehicles are braking, indicated by their acceleration values $a_L^{(i)}$ and $a_F^{(i)}$ being less than zero.
\item \textbf{Safety-Critical Event Detection}: A safety-critical event is identified if the DSS value is less than zero, indicating that the following vehicle may not maintain a safe distance if both vehicles continue to brake at their current rates.
\item \textbf{Recording Safety-Critical Events}: The time of the safety-critical event $t_C^{(i,j)}$ is recorded. If this is the first safety-critical event in the time series (evaluated through the $b_{C,1st}$ flag), the algorithm records the time of this first event $t_{C,1st}^{(i)}$ and sets the flag to True, indicating that a first occurrence has been identified.
\item \textbf{Invalid DSS Calculations}: If the condition for braking is not met, the DSS value for the data point is marked as NaN (Not a Number), indicating no valid safety assessment was performed due to the absence of braking by either vehicle.
\end{enumerate}

\section{Validation}
\begin{figure}[H]
\centering
    \includegraphics[scale=0.82]{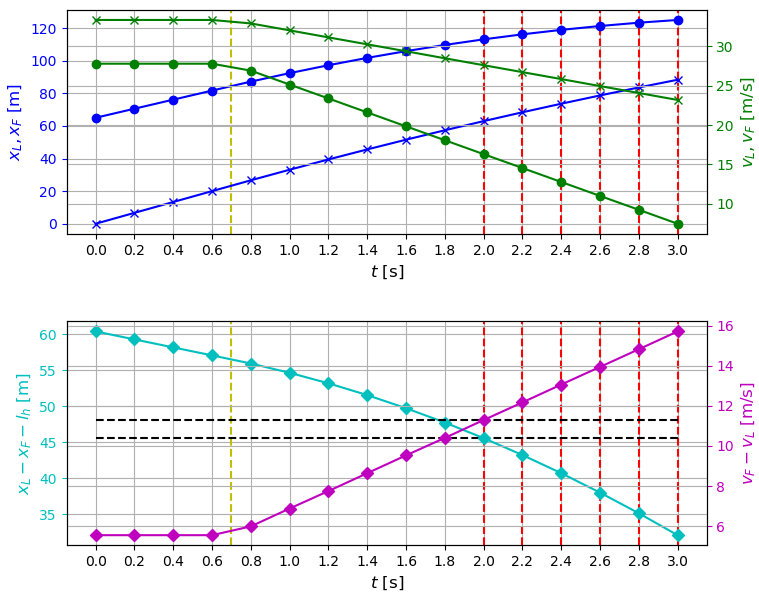}
    \caption{Validation: Follow-Up Drive (Illustration)}
    \label{Valid2}
\end{figure}

In this chapter, we present a comprehensive validation of the follow-up drive, as modeled in Algorithms \ref{algo:pseudo1} and \ref{algo:pseudo2}. The core of this validation is encapsulated in Figure \ref{Valid2}, which is composed of two distinct subplots.

The first subplot, positioned at the top, shows the absolute position and absolute velocity of the vehicles following each other. This representation is pivotal in illustrating the dynamic interaction and relative motion of the vehicles over time. The second subplot, located below, the focus shifts towards a more intricate aspect: the effective distance between the vehicles and their relative velocity. This subplot offers an insight into the spatiotemporal dynamics of the vehicles, with a specific focus on safety-related aspects.

A key assumption in our model is a uniform reaction time of 0.7 seconds for both vehicles. This parameter is graphically represented by a yellow line in the plots, serving as a visual anchor for interpreting the vehicles' response behavior. The time series follow a consistent temporal range, ranging from 0 to 3 seconds, with discrete intervals of 0.2 seconds. This level of granularity results in a dataset containing 16 pivotal data points, each presenting a snapshot of the system's state at a specific moment.

Critical moments within this time series, identified as safety-critical by DSS, are marked in red. These moments are primarily occurring between 2.0 and 3.0 seconds and are closely associated with a noticeable decrease in effective distance and a significant relative velocity difference, with the trailing vehicle exhibiting a significant speed excess relative to the leading vehicle.

A fundamental assumption in our analysis is the maximal deceleration capacity of the vehicles, quantified as $\mu \, g \approx 8.829$ m/s². This parameter is not merely a theoretical construct but a reflection of real-world vehicular capabilities, underscoring the realism and applicability of our model.

To gain a deeper insight into the specific DSS values, Table \ref{TabDSS2} offers a comprehensive breakdown of these values for the multivariate time series. This table offers a detailed examination of how far each time value deviates from the zero-meter DSS threshold, providing information about whether the corresponding time value is categorized as safety-critical or not.

The overarching objective of this validation is to underscore the importance of quantitative metrics in evaluating safety in driving scenarios. Through the utilization of metrics such as DSS, we gain invaluable insights into the dynamics of safety-critical situations. 

\begin{table}[h!]
\centering
\begin{adjustbox}{width=1\textwidth}
\begin{tabular}{|c|c|c|c|c|c|c|c|c|c|c|c|c|c|c|c|c|}
\hline
t [s] & 0.0 & 0.2 & 0.4 & 0.6 & 0.8 & 1.0 & 1.2 & 1.4 & 1.6 & 1.8 & 2.0 & 2.2 & 2.4 & 2.6 & 2.8 & 3.0 \\
\hline
DSS [m] & 17.86 & 16.75 & 15.64 & 14.53 & 12.63 & 9.98 & 7.49 & 5.02 & 2.63 & 0.40 & -1.80 & -3.91 & -5.93 & -7.83 & -9.66 & -11.41 \\
\hline
\end{tabular}
\end{adjustbox}
\vspace{0.1cm}
\caption{Validation: Follow-Up Drive (DSS Values)}
\label{TabDSS2}
\end{table}
\vspace{-0.5cm}
\section{Conclusion}
Our research represents a pivotal advancement in the field of road safety, particularly through the integration of intelligent driving functions tailored for safety-critical driving scenarios. 

Confronted with the challenge of limited testing data, we have developed an innovative methodology that significantly broadens the scope of testable safety-critical driving scenarios.

The core of our approach is the generation of realistic driving scenarios, represented as multivariate time series. These scenarios, carefully crafted to reflect realistic driving dynamics and temporal considerations, enable the simulation of varied vehicle interactions under diverse conditions, including the incorporation of crucial driver reaction times. This approach has allowed us to create a comprehensive range of driving behaviors within realistic dynamic parameters.

Our methodology's effectiveness is exemplified in a case study focusing on the follow-up drive movement of two vehicles, using kinematic models. This case study has proven instrumental in increasing the available test data for evaluating driver assistance systems and autonomous driving algorithms and technologies.

A key aspect of our research is the utilization of robust safety evaluation metrics, such as the Difference Space Stopping (DSS). Through the utilization of these metrics, we've successfully evaluated the safety significance of our generated driving scenarios, differentiating between those that are safety-critical and those that are not. This distinction is crucial for enhancing the validation process of driver assistance systems and autonomous vehicle functions across various driving scenarios, including potential edge cases.

\section{Appendix}
The two algorithms, Algorithm \ref{algo:pseudo1} and Algorithm \ref{algo:pseudo2}, are used for distinct purposes. The first algorithm is responsible for computing time series to represent follow-up driving scenarios, as detailed in section \ref{sec:algorithmEqMotion}. The second algorithm is dedicated to conducting safety assessments using DSS, as explained in section \ref{sec:algorithmDSS}. Both algorithms rely on multiple parameters, which are listed below.

\begin{algorithm}
  \caption{Parameters according to time series representing follow-up driving scenarios}
  \begin{algorithmic}[1]
    \State $N \subseteq \{1,2,\hdots,N_p\}$ \Comment{Total number of multivariate time series} 
    \State $n \subseteq \{1,2,\hdots,n_p\}$ \Comment{Total number of data points per multivariate time series} 
    \State $t_0 := t_{min} = 0 \, \left[s\right]$ \Comment{Initial time value of the uniform time vector} 
    \State $\Delta t = 0.2 \, \left[s\right]$ \Comment{Time step of the uniform time vector} 
    \State $l_V = 4.6 \, \left[m\right]$ \Comment{Vehicle length (applies equally to both vehicles)}
    \State $\mu_{a,L} := a_{min} := \mu \, g \approx 8.829 \, \left[m/s^2\right]$ \Comment{Expectation value (Acceleration) / Vehicle (L)}
    \State $\sigma_{a,L} := 1 \, \left[m/s^2\right]$ \Comment{Standard deviation (Acceleration) / Vehicle (L)}
    \State $\mu_{a,F} := a_{min} := \mu \, g \approx 8.829 \, \left[m/s^2\right]$ \Comment{Expectation value (Acceleration) / Vehicle (F)}
    \State $\sigma_{a,F} := 1 \, \left[m/s^2\right]$ \Comment{Standard deviation (Acceleration) / Vehicle (F)}
    \State $\mu_{x_L,0} = 65 \, \left[m\right]$ \Comment{Expectation value (Initial Position) / Vehicle (L)}
    \State $\sigma_{x_L,0} = 3 \, \left[m\right]$ \Comment{Standard deviation (Initial Position) / Vehicle (L)}
    \State $\mu_{v_L,0} = 27.78 \, \left[m/s\right]$ \Comment{Expectation value (Initial Velocity) / Vehicle (L)}
    \State $\sigma_{v_L,0} = 1 \, \left[m/s\right]$ \Comment{Standard deviation (Initial Velocity) / Vehicle (L)}
    \State $\mu_{x_F,0} = 0 \, \left[m\right]$ \Comment{Expectation value (Initial Position) / Vehicle (F)}
    \State $\sigma_{x_F,0} = 3 \, \left[m\right]$ \Comment{Standard deviation (Initial Position) / Vehicle (F)}
    \State $\mu_{v_F,0} = 33.33 \, \left[m/s\right]$ \Comment{Expectation value (Initial Velocity) / Vehicle (F)}
    \State $\sigma_{v_F,0} = 1 \, \left[m/s\right]$ \Comment{Standard deviation (Initial Velocity) / Vehicle (F)}
    \State $x_{L,0} := \mathcal{N}(\mu_{x_L,0},\sigma_{x_L,0}) \, \left[m\right]$ \Comment{Initial Position / Vehicle (L)} 
    \State $v_{L,0} := \mathcal{N}(\mu_{v_L,0},\sigma_{v_L,0}) \, \left[m/s\right]$ \Comment{Initial Velocity / Vehicle (L)}
    \State $x_{F,0} := \mathcal{N}(\mu_{x_F,0},\sigma_{x_F,0}) \, \left[m\right]$ \Comment{Initial Position / Vehicle (F)}
    \State $v_{F,0} := \mathcal{N}(\mu_{v_F,0},\sigma_{v_F,0}) \, \left[m/s\right]$ \Comment{Initial Velocity / Vehicle (F)}
  \end{algorithmic}
  \label{algo:pseudo}
\end{algorithm}

\end{document}